\documentclass[twocolumn, switch]{article} 

\usepackage{adjustbox}
\usepackage{preprint}
\usepackage{algorithm}
\usepackage{algorithmic}
\usepackage{amsmath, amsthm, amssymb, amsfonts}
\usepackage{multirow}
\usepackage[numbers,square]{natbib}
\usepackage{subfig}
\usepackage[utf8]{inputenc}	
\usepackage[T1]{fontenc}	
\usepackage{xcolor}		
\usepackage[colorlinks = true,
            linkcolor = purple,
            urlcolor  = blue,
            citecolor = cyan,
            anchorcolor = black]{hyperref}	
\usepackage{booktabs} 		
\usepackage{nicefrac}		
\usepackage{microtype}		
\usepackage{lineno}		
\usepackage{float}			

\usepackage{lipsum}		

\usepackage{newfloat}
\DeclareFloatingEnvironment[name={Supplementary Figure}]{suppfigure}
\usepackage{sidecap}
\sidecaptionvpos{figure}{c}

\usepackage{titlesec}
\titlespacing\section{0pt}{12pt plus 3pt minus 3pt}{1pt plus 1pt minus 1pt}
\titlespacing\subsection{0pt}{10pt plus 3pt minus 3pt}{1pt plus 1pt minus 1pt}
\titlespacing\subsubsection{0pt}{8pt plus 3pt minus 3pt}{1pt plus 1pt minus 1pt}

\title{Enhancing LiDAR Point Features with Foundation Model Priors for 3D Object Detection}

\usepackage{eso-pic}
\usepackage{tikz}
\usepackage{xcolor}
\PassOptionsToPackage{colorlinks=true,linkcolor=gray,urlcolor=gray}{hyperref}

\newcommand{\AddMyWatermarks}{%
  \begin{tikzpicture}[remember picture, overlay]
    \node[color=gray!90, scale=1] at ([xshift=0in,yshift=-5in]current page.center) {%
      This is the author's accepted manuscript. The final version will appear in ITSC 2025, © IEEE 2025.%
    };
  \end{tikzpicture}%
}

\AddToShipoutPictureBG*{\AddMyWatermarks}

\usepackage{titling}
\usepackage{orcidlink}
\usepackage{footmisc}
\newcommand{\Author}[2]{
  \textbf{#1}\textsuperscript{#2}%
}

\author{
  \Author{Yujian Mo}{1}\and
  \Author{Yan Wu}{1,*}\and
  \Author{Junqiao Zhao}{1}\and
  \Author{Jijun Wang}{1}\and
  \Author{Yinghao Hu}{1}\and
  \Author{Jun Yan}{2}\and
}

\date{%
  \textsuperscript{1}School of Computer Science and Technology, Tongji University, Shanghai 201804, China\\
  \textsuperscript{2}School of Electronics and Information Engineering, Tongji University, Shanghai 201804, China\\[1em]
  \footnotesize \textbf{*Corresponding author:} Yan Wu\\
}

\begin{document}

\twocolumn[ 
  \begin{@twocolumnfalse} 

\maketitle
\thispagestyle{empty}

\begin{abstract}
Recent advances in foundation models have opened up new possibilities for enhancing 3D perception. In particular, DepthAnything offers dense and reliable geometric priors from monocular RGB images, which can complement sparse LiDAR data in autonomous driving scenarios. However, such priors remain underutilized in LiDAR-based 3D object detection. In this paper, we address the limited expressiveness of raw LiDAR point features, especially the weak discriminative capability of the reflectance attribute, by introducing depth priors predicted by DepthAnything. These priors are fused with the original LiDAR attributes to enrich each point's representation. To leverage the enhanced point features, we propose a point-wise feature extraction module. Then, a Dual-Path RoI feature extraction framework is employed, comprising a voxel-based branch for global semantic context and a point-based branch for fine-grained structural details. To effectively integrate the complementary RoI features, we introduce a bidirectional gated RoI feature fusion module that balances global and local cues. Extensive experiments on the KITTI benchmark show that our method consistently improves detection accuracy, demonstrating the value of incorporating visual foundation model priors into LiDAR-based 3D object detection. 
\end{abstract}
\vspace{0.35cm}

  \end{@twocolumnfalse} 
] 


\section{INTRODUCTION}

Recent years have witnessed the emergence of foundation models, which are large-scale models pre-trained on diverse and massive datasets, demonstrating strong generalization abilities across various downstream tasks. These models, including Vision Transformers, CLIP \cite{_A_Radford___LTVMFNLS___}, and Segment Anything Model (SAM) \cite{_Y_Xiong___ELMIPfESA___,_A_Kirillov___SA___}, have revolutionized fields like image classification, semantic segmentation, and depth estimation by offering strong semantic and geometric priors with minimal task-specific supervision \cite{Mo2022626}.

The field of computer vision and natural language processing is undergoing a paradigm shift with the emergence of these foundation models, which can provide powerful prior information for various downstream tasks, such as object detection and semantic segmentation \cite{_L_Yang_B_Kang__DAUtPoLSUD___,_L_Yang_B_Kang__DAV__2024_,_R_Bommasani___OtOaRoFM___,wang2025collaborative}. 
In 3D object detection, LiDAR point clouds have become a dominant sensing modality due to their accurate 3D spatial measurements \cite{_S_Shi___PRPVFSAf3OD___,_Y_Yan___SSECD___,_G_Zhang___VMGFSSMfPCb3OD___}.
Compared to 2D images, LiDAR directly captures critical geometric properties such as boundaries, contours, and volumetric shapes. 
These properties contribute to more robust and accurate 3D detection performance.

\begin{figure}[!t]
  \centering

  \subfloat[]{\includegraphics[width=0.44\textwidth]{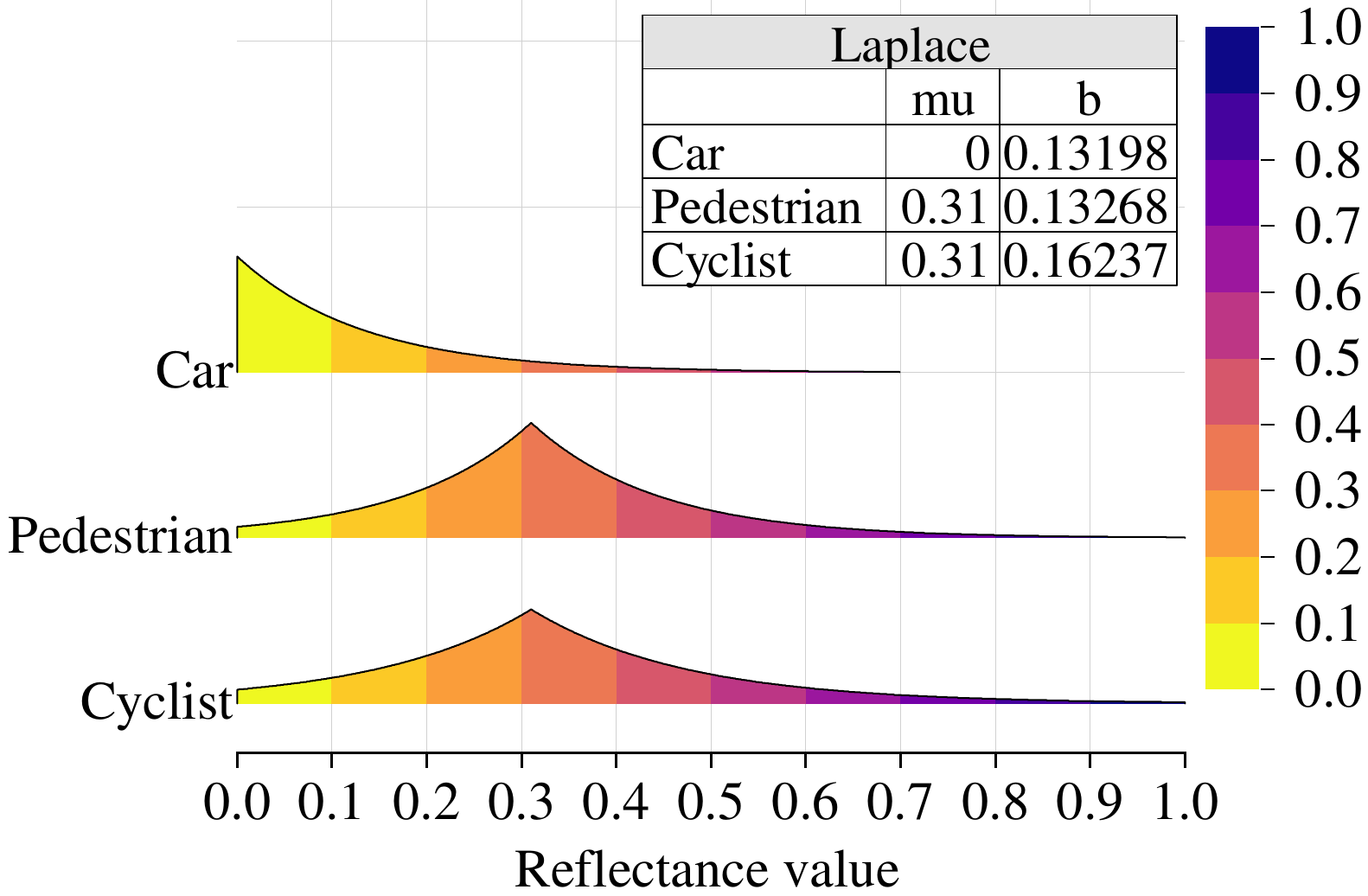}}

  \subfloat[]{\includegraphics[width=0.46\textwidth]{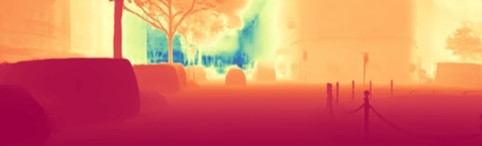}}
  
  
    \caption{(a) Distribution of reflectance values for Car, Pedestrian, and Cyclist categories. The Car is mainly concentrated in [0, 0.1], while Pedestrian and Cyclist show substantial overlap, indicating limited discriminative power of reflectance values. 
    (b) Depth map predicted by DepthAnything V2, showing clear separation of objects at different distances. 
    }
    \label{fig:multirow-figure}
\end{figure}

However, raw LiDAR data exhibits inherent limitations. In datasets such as KITTI, each LiDAR point is described by $(x, y, z, r)$, where $r$ represents the reflectance \cite{_T_Yin___MVP3D___}. 
While spatial coordinates $(x, y, z)$ provide strong geometric cues, the reflectance $r$ is affected by factors such as surface material and incident angle, leading to weak discriminative capability. 
To further illustrate this issue, we analyze the distribution of reflectance values across different object categories. 
As illustrated in Fig. \ref{fig:multirow-figure}(a), the reflectance values of the Car category are predominantly concentrated in [0, 0.1], whereas the Pedestrian and Cyclist categories exhibit significant overlap. 
This observation highlights the limited utility of reflectance for reliably distinguishing between object types in 3D detection tasks.

Classical methods have addressed this issue by incorporating depth priors through depth completion \cite{_H_Wu___VSCfM3OD___}. 
However, these methods often require dataset-specific pretraining or fine-tuning, restricting generalization \cite{_J_Tang_F_Tian__BPNfDC___,_Y_Wang_G_Zhang__IDCvDFU___}. 
In contrast, foundation models offer strong generalization capabilities out of the box. For example, DepthAnything \cite{_L_Yang_B_Kang__DAUtPoLSUD___} and its successor DepthAnything V2 \cite{_L_Yang_B_Kang__DAV__2024_} provide dense depth predictions from single RGB images without dataset-specific fine-tuning.
Rather than serving as a replacement for LiDAR's accurate spatial measurements, the depth prior predicted by DepthAnything is treated as a discriminative 3D cue that enhances the geometric separability of LiDAR points in feature space.
As illustrated in Fig. \ref{fig:multirow-figure}(b), DepthAnything V2 produces a dense and perceptually clear depth map, where objects at varying distances are distinctly separated, providing more informative and reliable geometric cues than the raw reflectance values.

Motivated by these insights, we propose a Dual-Path RoI feature extraction and fusion framework for 3D object detection. 
We first enhance raw LiDAR point representations by incorporating depth priors predicted by DepthAnything V2, alongside the original spatial coordinates and reflectance, forming enriched five-dimensional point features. 
These enriched points are then processed by a point-wise feature extractor, PointGFE, which explicitly incorporates the depth prior into local geometric encoding.
RoI features are then extracted through two complementary paths: a voxel-based branch that captures global semantic context via RoI Grid Pooling \cite{_J_Deng___VRCTHPVb3OD___}, and a point-based branch that preserves fine-grained geometric structures using RoI Aware Pooling \cite{_S_Shi___FPtP3ODFPCWPAaPAN___}. 
To effectively integrate these complementary RoI features, we further introduce a bidirectional gated fusion module that dynamically balances global and local cues via adaptive attention-based gating for RoI representation refinement.
Extensive experiments on the KITTI benchmark validate the effectiveness of our method.

Our contributions can be summarized as follows:
\begin{itemize}
\item We propose a depth prior augmentation strategy using foundation models to enrich LiDAR points with geometric prior cues, enhancing LiDAR point representation without requiring dataset-specific adaptation.
\item We design a Dual-Path RoI feature extraction framework that combines voxel-based global context via RoI Grid Pooling and fine-grained local geometry via PointGFE and RoI Aware Pooling, followed by a bidirectional gated fusion module for adaptive feature fusion.
\item Extensive experiments on KITTI demonstrate consistent performance improvements across Car, Pedestrian, and Cyclist categories, verifying our method's effectiveness.
\end{itemize}

\section{RELATED WORK}
\subsection{Voxel-based 3D Object Detection}
Voxel-based methods have become a dominant paradigm in 3D object detection due to their structured data representation and computational efficiency.
SECOND \cite{_Y_Yan___SSECD___} improves upon VoxelNet \cite{_Y_Zhou___VEtELfPCB3OD___} by introducing a GPU-accelerated sparse convolution algorithm, which significantly speeds up voxel-based feature extraction while maintaining accuracy. It further proposes ground-truth sampling for data augmentation and a tailored angle loss to mitigate orientation ambiguities in box regression.
Voxel R-CNN \cite{_J_Deng___VRCTHPVb3OD___} extends this paradigm by preserving fine-grained 3D structure. It refines RoI proposals using 3D voxel features through a two-stage pipeline, improving localization precision by capturing neighborhood-aware context.

To overcome the inherent limitations of raw LiDAR features, especially the low discriminative power of reflectance, recent studies have explored ways to enrich point-level information.
PointAugmenting  \cite{_C_Wang___PCMAf3OD___} supplements each LiDAR point with auxiliary features from external models, while PointPainting \cite{_S_Vora___PSFf3OD___} maps LiDAR points onto semantic segmentation maps derived from RGB images to inject semantic priors. 
These approaches have shown notable improvements, particularly for detecting small or distant objects.

Inspired by these advancements, we introduce a novel strategy that leverages depth priors predicted by foundation models to enrich LiDAR point features, addressing the limited expressiveness of reflectance attributes.

\subsection{Foundation Models}
Recent advances in foundation models have significantly broadened the capabilities of vision systems across a variety of tasks. The SAM introduces a universal segmentation framework that generalizes across diverse scenes and object categories. Trained on over a billion masks, SAM enables high-quality instance segmentation based on prompts such as points, boxes, or free-form text, offering dense spatial representations useful for image-level understanding \cite{_Y_Xiong___ELMIPfESA___,_A_Kirillov___SA___}.

In the domain of monocular depth estimation, DepthAnything and its successor DepthAnything V2 leverage strong vision backbones to predict dense depth maps from single RGB images \cite{_L_Yang_B_Kang__DAUtPoLSUD___,_L_Yang_B_Kang__DAV__2024_}. 
By training on heterogeneous large-scale datasets, these models exhibit strong generalization across indoor and outdoor environments. 

Additionally, contrastive vision-language models such as CLIP \cite{_A_Radford___LTVMFNLS___}, and its derivatives, including BLIP \cite{_J_Li___BBLIPtfUVLUaG___} and GLIP \cite{_L_Li___GLIPt___}, align visual and textual modalities in a shared embedding space. 

Building upon these developments, our work explores the integration of depth priors from foundation models into LiDAR point features. By fusing predicted depth information with spatial coordinates and reflectance attributes, we aim to improve the discriminative power of point representations and enhance downstream 3D object detection performance.

\begin{figure}[!t]
\centerline{\includegraphics[width=1\linewidth]{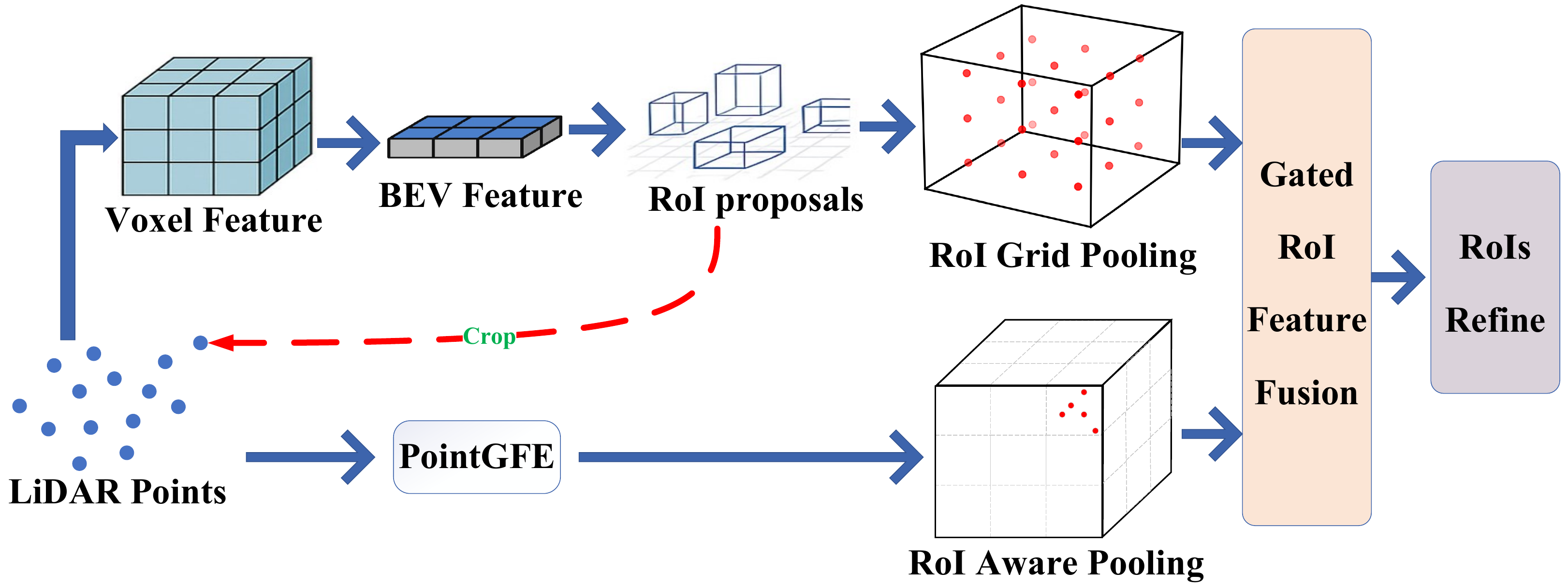}}
\caption{Overview of the proposed 3D object detection framework. We first enhance each LiDAR point by fusing its original attributes with depth priors predicted by DepthAnything V2. The enriched points are processed through two parallel branches: (top) a voxel-based branch extracts global semantic features via 3D sparse convolution and RoI Grid Pooling; (bottom) a point-based branch captures fine-grained local structures through PointGFE and RoI Aware Pooling. The two RoI features are dynamically integrated using a gated fusion module to refine RoI proposals.}
\label{fig:pipeline}
\end{figure}

\section{METHOD}
As illustrated in Fig. \ref{fig:pipeline}, we propose a 3D object detection framework that enhances LiDAR point features by integrating depth priors predicted by foundation models. 
Specifically, we fuse depth priors with original LiDAR attributes to obtain enriched point features. 
Building on these, we design two complementary RoI feature extraction branches: a voxel-based branch using RoI Grid Pooling to capture global semantic context, and a point-based branch that preserves fine-grained geometry via PointGFE and RoI Aware Pooling.
To effectively integrate the complementary information from both branches, we introduce a bidirectional gated fusion module that dynamically balances global and local cues. 
This unified framework produces more discriminative and robust RoI feature representations.

\subsection{Depth Prior Augmentation for LiDAR Points}
Although LiDAR provides accurate 3D spatial coordinates, its reflectance attribute $r$ is highly sensitive to surface materials, sensor distance, and incidence angles, resulting in weakly discriminative features.
To address this limitation, we propose a depth prior augmentation strategy to enrich each LiDAR point.

Using the known intrinsic and extrinsic calibration between the LiDAR sensor and the camera, each LiDAR point $(x, y, z)$ is projected onto the image plane to obtain its corresponding 2D pixel coordinates.
The predicted depth value $d_{\text{DA}}$ from DepthAnything V2 output is then sampled at these locations. 
This predicted depth prior serves as a complementary geometric cue, independent of the original LiDAR measurements, and offers an additional perspective for describing the 3D scene structure.

The retrieved depth prior is concatenated with the original LiDAR attributes to form an enhanced five-dimensional representation $(x, y, z, r, d_{\text{DA}})$ for each point. 
These enriched points are voxelized and fed into a 3D sparse convolutional backbone to extract structured voxel-wise representations. 
These representations serve as the foundation for downstream 3D object detection tasks, including proposal generation and region refinement.

\subsection{Dual-Path RoI Feature Extraction}
To enhance the quality of RoI features, we propose a Dual-Path RoI feature extraction strategy that combines voxel-based and point-based representations.

\subsubsection{Voxel-based RoI Feature Extraction}

To fully exploit the global contextual information encoded by the 3D sparse convolutional backbone, we adopt a voxel-based RoI feature extraction strategy based on RoI Grid Pooling.
Given a set of RoIs generated by the one-stage of Voxel R-CNN \cite{_J_Deng___VRCTHPVb3OD___}, each RoI uniformly generates $n \times n \times n$ grid points. 

Each grid point queries neighboring voxel features from the 3D backbone feature map using a ball query mechanism.
The features of all grid points are then concatenated to form a fixed-size RoI representation that is robust to variations in object scale and point density. 

These voxel-based RoI features effectively preserve the coarse spatial structures and semantic information captured during voxelization, providing strong global priors that are crucial for accurate object classification and bounding box refinement.

\begin{figure}[!t]
\centerline{\includegraphics[width=1\linewidth]{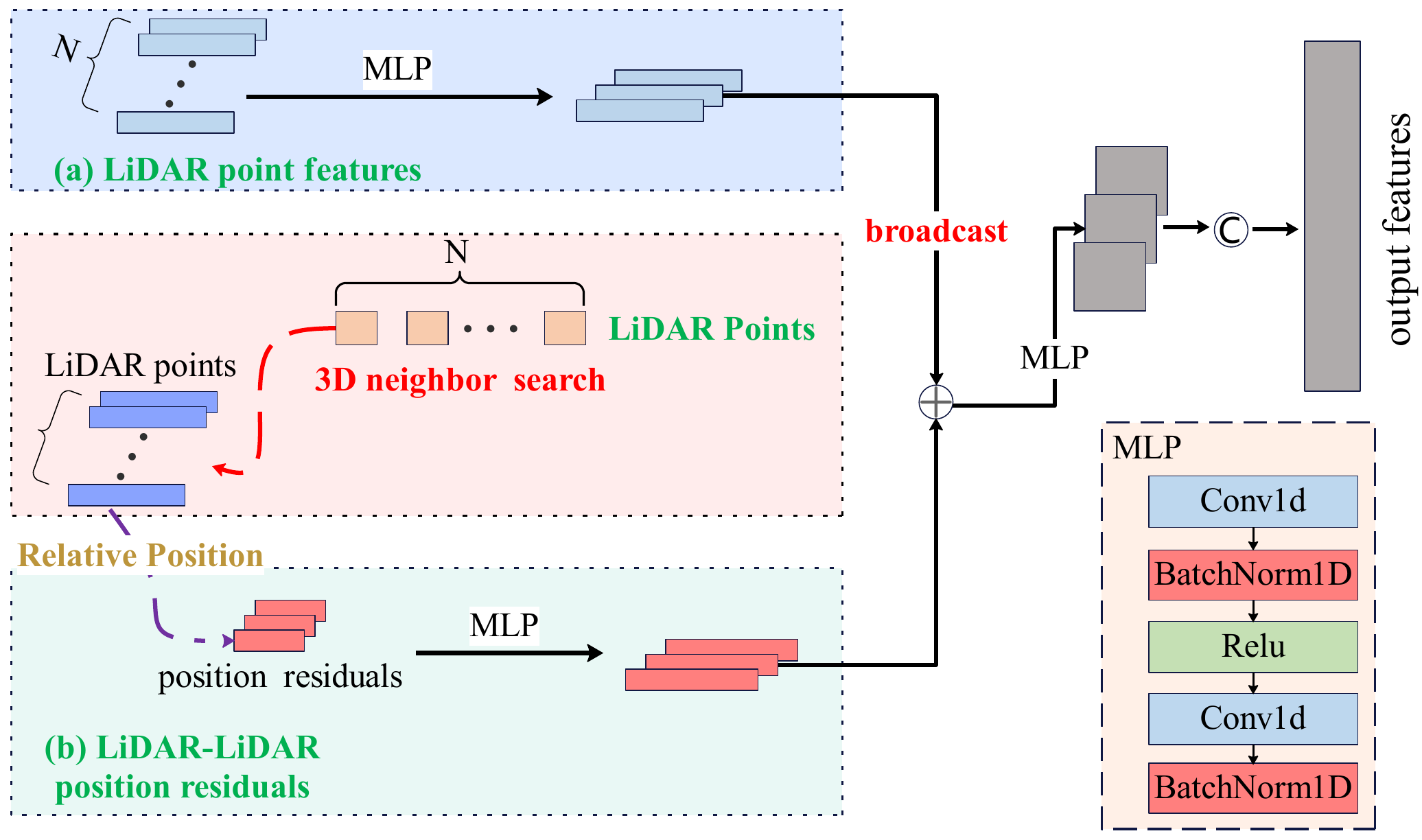}}
\caption{Illustration of the proposed PointGFE module. Each LiDAR point is enriched with a depth prior, which is explicitly used as one input dimension. For each LiDAR point, neighboring points within a predefined radius are identified using a ball query. The relative positions between the LiDAR point and its neighbors are computed and fused with the original point-wise features.}
\label{fig:pointGFE}
\end{figure}


\subsubsection{Point-based RoI Feature Extraction}
To complement voxel-based RoI features with fine-grained geometric details, we introduce a point-based RoI feature extraction strategy.
Given a set of RoIs produced by the one-stage of Voxel R-CNN, we first crop the LiDAR points falling inside each RoI. To ensure adequate point coverage, each RoI is slightly enlarged before point assignment.

For each point within a RoI, we apply a GPU-accelerated ball query to identify neighboring LiDAR points within a predefined radius $r$. 
The relative positional offsets between each point and its neighbors are computed to encode the local geometric structure.
To eliminate the effect of differing RoI orientations, all local structures are transformed into a canonical coordinate frame via rotation normalization.

To effectively capture fine-grained local geometry from the enriched LiDAR points, we introduce a dedicated point feature extraction module, termed PointGFE, as illustrated in Fig. \ref{fig:pointGFE}.
Within this module, point-wise features are broadcast and fused with their corresponding local geometric features, followed by a multi-layer perceptron (MLP) for feature transformation and refinement.
Notably, the depth prior predicted by DepthAnything is simply concatenated with each LiDAR point's attributes. Through the multi-stage feature extraction in PointGFE, the network implicitly learns to utilize this additional input to enhance local feature representation.


We stack three successive PointGFE modules, each progressively refining the point-wise representation.
The output features from all three stages are then concatenated to form the final point-wise embedding, which captures fine-grained geometric details.

After extracting enriched point-wise features using PointGFE, we aggregate them into structured RoI representations via RoI Aware Pooling.
Specifically, each RoI is uniformly divided into $m \times m \times m$ sub-voxels.
LiDAR points within the RoI are assigned to corresponding sub-voxels based on their spatial locations.
For each sub-voxel, we apply a max-pooling operation over the features of the points falling within it, thereby obtaining a compact feature representation. The resulting sub-voxel features are assembled into a 3D RoI feature volume.

To align with the voxel-based RoI features obtained via RoI Grid Pooling, we apply a sparse 3D convolution to downsample the $m \times m \times m$ feature volume to a fixed resolution of $n \times n \times n$.
This alignment ensures that both voxel-based and point-based RoI features share a consistent spatial layout, facilitating effective fusion in the subsequent bidirectional gated fusion module.

\subsection{Bidirectional Gated RoI Feature Fusion}
After extracting voxel-based and point-based RoI features through their respective paths, we employ a bidirectional gated fusion module to integrate the complementary information from the two RoI features.

\begin{figure}[!t]
\centerline{\includegraphics[width=0.95\linewidth]{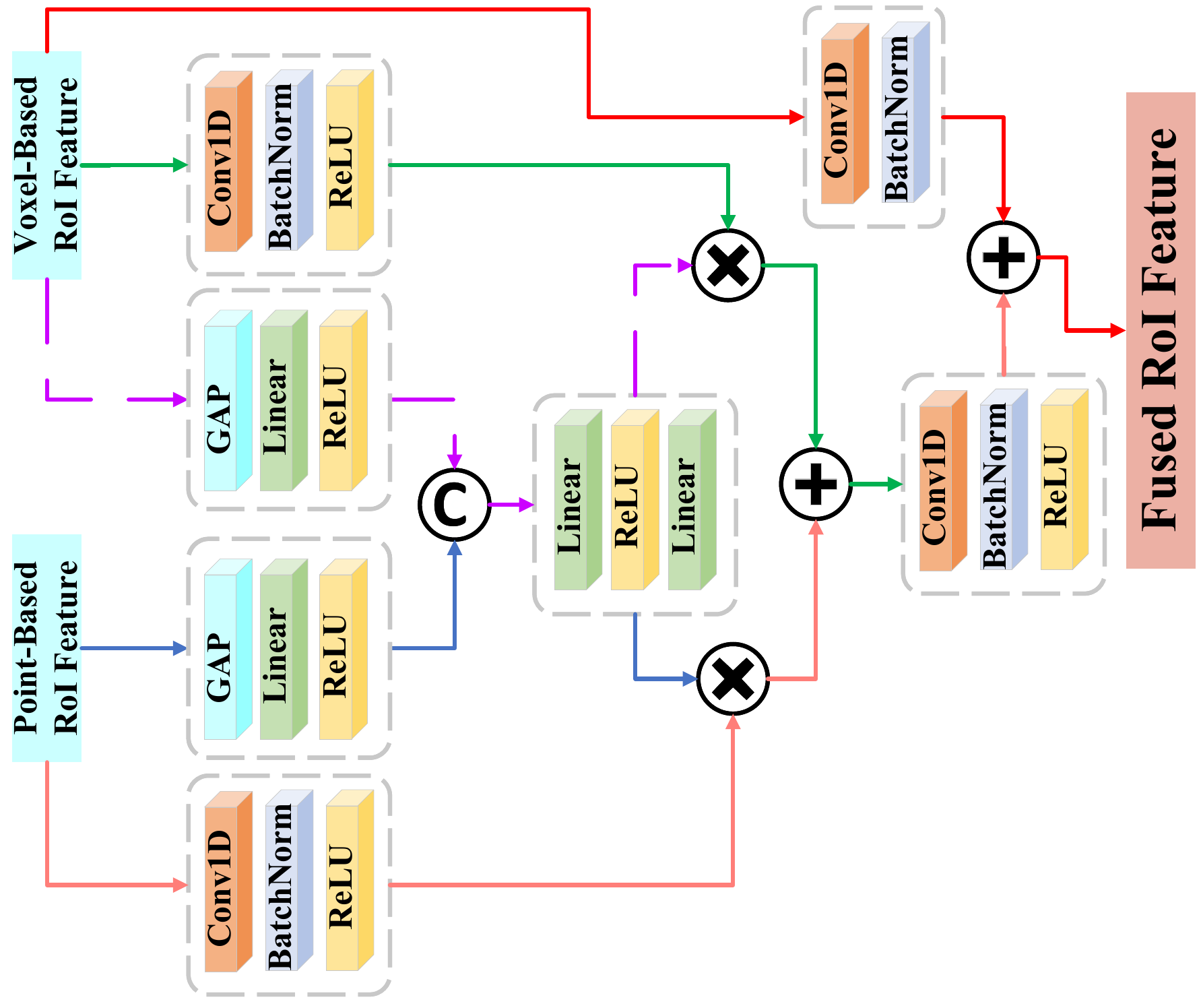}}
\caption{Overview of the bidirectional gated fusion module. RoI features from the voxel-based and point-based branches are first compressed using global average pooling (GAP), followed by a linear layer and ReLU activation to unify their channel dimensions. These compressed features are then fed into a dual-branch gating network, implemented with a Linear–ReLU–Linear structure, to predict adaptive weights that dynamically balance their contributions at each spatial location. The weighted features are fused via element-wise summation, followed by refinement via convolutional layers and a residual connection.}
\label{fig:bgrf}
\end{figure}

The voxel-based RoI features, obtained through RoI Grid Pooling, encode coarse global spatial structures and semantic context. In parallel, point-based RoI features, aggregated via PointGFE and RoI Aware Pooling, capture fine-grained local geometric details. 
Before fusion, both types of RoI features are spatially aligned and normalized to the same grid resolution to ensure accurate correspondence across spatial locations. The fusion architecture is illustrated in Fig. \ref{fig:bgrf}.

Inspired by the Cascade Attention mechanism \cite{_B_Fan___HHCPVFf3OD___,_A_Liu___CRCSANfHQ3DODFLPC___,_H_Wu___CACANf3DODFLPC___}, we stack three stages of bidirectional gated fusion module to progressively integrate voxel-based and point-based RoI features.
At each stage, learned attention cues guide a pair of gating branches to adaptively balance contributions from both RoI feature streams at each spatial location, enabling the dynamic fusion of global semantics and local geometry.
Moreover, by fusing point-based and voxel-based RoI features in a gated, bidirectional manner, the Bidirectional Gated RoI Fusion module (BGRF) allows depth-enriched point features to compensate for potential prior loss caused by voxelization.

Each fusion stage is followed by a dedicated RoI head that independently refines proposals. Higher stages further refine the outputs of earlier stages, enabling deep interaction between complementary features. 
Final predictions are obtained by averaging the outputs from all three stages.
This cascade fusion strategy facilitates progressive feature alignment and hierarchical integration, yielding more discriminative RoI representations and improved detection performance.

\section{EXPERIMENTAL RESULTS}
\subsection{Dataset}
The KITTI dataset \cite{Geiger_Lenz_AwrfadTKvbs_} is one of the most widely used 3D object detection datasets. It consists of 7,481 training samples and 7,518 testing samples. 
Following standard practice, we split the training set into 3,712 samples for training and 3,769 samples for validation.

KITTI evaluates 3D object detection performance on the Car, Pedestrian, and Cyclist categories. Each category is further divided into \textit{Easy}, Moderate (\textit{Mod.}), and \textit{Hard} difficulty levels based on bounding box height, occlusion, and truncation.

\subsection{Implementation Details}
We implement our method based on the OpenPCDet \cite{openpcdet2020}. For local geometric encoding, we use a ball query with a radius $r$ of 0.8 to select 9 neighboring LiDAR points for each query point. The relative positions between each LiDAR point and its neighbors are computed and encoded to capture fine-grained local geometry.

Our framework is trained on two NVIDIA RTX 3090 GPUs. For the point-based RoI feature extraction, we adopt RoI Aware Pooling with $m = 12$, resulting in $12 \times 12 \times 12$ sub-voxels per RoI. 
To ensure feature alignment with the voxel-based RoI Grid Pooling (which outputs features at $6 \times 6 \times 6 $), we apply a sparse 3D convolution to downsample the RoI-aware pooled features to the same spatial resolution.

\subsection{Main Results}
We evaluate our proposed method on both the KITTI test and validation sets across three object categories: Car, Pedestrian, and Cyclist. 
As shown in Tab. \ref{tab:test}, on the official test set, our approach achieves competitive performance, with 90.98 / 82.35 / 77.26 for Car, 49.04 / 41.85 / 38.29 for Pedestrian, and 79.96 / 66.47 / 58.47 for Cyclist under the \textit{Easy} / \textit{Mod}. / \textit{Hard} difficulty levels, respectively. 
The overall mean AP reaches 64.96, which is on par with a variety of strong LiDAR-based detectors such as PV-RCNN \cite{_S_Shi___PRPVFSAf3OD___} and PG-RCNN \cite{_I_Koo___PRSSPGf3OD___}. Notably, our method achieves comparable results to multi-modal methods like PointPainting \cite{_S_Vora___PSFf3OD___} and DFAF3D \cite{_Q_Tang___DAdfaafss3dfpc___}.

\begin{table*}[t]
\centering
\caption{Quantitative detection performance for multi-class detection on the KITTI test set using AP R40.}
\label{tab:kitti-test}
\begin{tabular}{l|ccc|ccc|ccc|c}
\toprule
\multirow{2}{*}{Method} & \multicolumn{3}{c|}{Car 3D(R40)} & \multicolumn{3}{c|}{Pedestrian 3D(R40)} & \multicolumn{3}{c|}{Cyclist 3D(R40)} & \textbf{MAP} \\
\cmidrule(r){2-4} \cmidrule(r){5-7} \cmidrule(r){8-10}
& \textit{Easy} & \textit{Mod}. & \textit{Hard} & Easy & \textit{Mod}. & \textit{Hard} & \textit{Easy} & \textit{Mod}. & \textit{Hard} & \\
\midrule
Voxel R-CNN \cite{_J_Deng___VRCTHPVb3OD___} & 90.90 & 81.62 & 77.06 & - & - & - & - & - & - & - \\
GD-MAE \cite{_H_Yang___GMGDfMPtoLPC___} & 88.14 & 79.03 & 73.55 & - & - & - & - & - & - & - \\
EOTL  \cite{_R_Yang___EOTLfRPDiAD___} & 79.97 & 69.13 & 58.57 & 48.65 & 40.11 & 35.99 & 75.20 & 58.96 & 50.41 & 57.44 \\
SeSame \cite{_O_Hayeon___SSE3ODwPWS__990_}  & 81.51 & 75.05 & 70.53 & 46.53 & 37.37 & 33.56 & 70.97 & 54.36 & 48.66 & 57.62 \\
HINTED \cite{_Q_Xia___HHIEDwMDFFfSS3OD__986_} & 84.00 & 74.13 & 67.03 & 47.33 & 37.75 & 34.10 & 76.21 & 63.01 & 55.85 & 59.93 \\
PointPainting \cite{_S_Vora___PSFf3OD___} & 82.11 & 71.70 & 67.08 & 50.32 & 40.97 & 37.87 & 77.63 & 63.78 & 55.89 & 60.82 \\
DFAF3D  \cite{_Q_Tang___DAdfaafss3dfpc___} & 88.59 & 79.37 & 72.21 & 47.58 & 40.99 & 37.65 & 82.09 & 65.86 & 59.02 & 63.71 \\
GraphAlign \cite{_Z_Song___GEAFAbGmfMM3OD___}  & 90.90 & 82.23 & 79.67 & 41.38 & 36.89 & 34.95 & 78.42 & 64.43 & 58.71 & 63.06 \\
PV-RCNN  \cite{_S_Shi___PRPVFSAf3OD___} & 90.25 & 81.43 & 76.82 & 52.17 & 43.29 & 40.29 & 78.60 & 63.71 & 57.65 & 64.91 \\
PG-RCNN \cite{_I_Koo___PRSSPGf3OD___} & 89.38 & 82.13 & 77.33 & 47.99 & 41.04 & 38.71 & 82.77 & 67.82 & 61.25 & 65.38 \\
\textbf{Ours} & \textbf{90.98} & \textbf{82.35} & \textbf{77.26} & \textbf{49.04} & \textbf{41.85} & \textbf{38.29} & \textbf{79.96} & \textbf{66.47} & \textbf{58.47} & \textbf{64.96} \\
\bottomrule
\end{tabular}
\label{tab:test}
\end{table*}

Rather than pursuing state-of-the-art performance, our primary goal is to validate the effectiveness of introducing depth priors predicted by foundation models into LiDAR-based 3D object detection. 
Through PointGFE and BGRF, the predicted depth is actively utilized throughout both feature extraction and fusion stages, leading to enhanced representation quality, particularly for sparse scenes and small-scale objects.

We further evaluate our method on the KITTI validation set for the Car category. Tab. \ref{tab:main_result} shows that our method achieves 93.12, 86.13, and 83.65 in the \textit{Easy} / \textit{Mod}. / \textit{Hard} difficulty levels, surpassing several strong voxel-based baselines, including Voxel R-CNN \cite{_J_Deng___VRCTHPVb3OD___} (92.38 / 85.29 / 82.86) and PV-RCNN \cite{_S_Shi___PRPVFSAf3OD___} (92.10 / 84.36 / 82.48). In terms of Bird's Eye View (BEV) detection, our method also achieves the highest AP across all difficulty levels.


\begin{table}[htbp]
\centering
\caption{Quantitative detection performance for Car detection on the KITTI val set using AP R40.}
\resizebox{\columnwidth}{!}{%
\begin{tabular}{lcccccc}
\toprule
\multirow{2}{*}{\textbf{Method}} & \multicolumn{3}{c}{\textbf{Car 3D (R40)}} & \multicolumn{3}{c}{\textbf{Car BEV (R40)}} \\
\cmidrule(lr){2-4} \cmidrule(lr){5-7}
 & \textit{Easy} & \textit{Mod}. & \textit{Hard} & \textit{Easy} & \textit{Mod}. & \textit{Hard} \\
\midrule
Voxel R-CNN  \cite{_J_Deng___VRCTHPVb3OD___} & 92.38 & 85.29 & 82.86 & 95.52 & 91.25 & 88.99 \\
PointPainting \cite{_S_Vora___PSFf3OD___} & -     & -     & -     & 90.05 & 87.51 & 86.66 \\
PV-RCNN  \cite{_S_Shi___PRPVFSAf3OD___}     & 92.10 & 84.36 & 82.48 & 95.76 & 91.11 & 88.93 \\
\textbf{Ours} & \textbf{93.12} & \textbf{86.13} & \textbf{83.65} & \textbf{96.05} & \textbf{91.99} & \textbf{89.71} \\
\bottomrule
\end{tabular}
}
\label{tab:main_result}
\end{table}

As shown in Tab. \ref{tab:multi_class}, integrating depth priors predicted by DepthAnything V2 leads to consistent performance gains for Pedestrian (+3.33) and Cyclist (+1.12) categories on the KITTI validation set.

For the Car category, where LiDAR already provides reliable geometric information, incorporating depth priors results in a slight performance drop of 1.05. 
This may be attributed to redundant information for potential misalignment between image-predicted depth and actual LiDAR geometry, particularly over large or reflective surfaces. 
These results suggest that while foundation model priors are particularly beneficial for detecting small or challenging objects, their integration into LiDAR-based pipelines for geometrically clear objects requires careful consideration.



\begin{table}[htbp]
\centering
\caption{Quantitative detection performance for multi-class detection on the KITTI val set using AP R40. We compare the performance w./w.o. depth priors.}
\label{tab:depth_prior_comparison}
\resizebox{\columnwidth}{!}{%
\begin{tabular}{cc|ccc|c}
\toprule
Method & w/o Depth Prior & \textit{Easy} & \textit{Mod}. & \textit{Hard} & mAP \\
\midrule
\multirow{2}{*}{Car} & $\times$ & 95.46 & 86.48 & 83.82 & 88.59 \\
                     & $\checkmark$ & 93.04 & 86.04 & 83.54 & 87.54 (-1.05) \\
\midrule
\multirow{2}{*}{Pedestrian} & $\times$ & 62.75 & 56.17 & 50.09 & 56.34 \\
                            & $\checkmark$ & 66.33 & 59.60 & 53.07 & 59.67 (+3.33) \\
\midrule
\multirow{2}{*}{Cyclist} & $\times$ & 87.74 & 66.83 & 62.27 & 72.28 \\
                         & $\checkmark$ & 90.41 & 67.27 & 62.53 & 73.40 (+1.12) \\
\bottomrule
\end{tabular}
}
\label{tab:multi_class}
\end{table}

\subsection{Ablation Study}
Tab. \ref{tab:abla} presents the ablation study on the KITTI validation set for the Car category, evaluating the individual and combined effects of two core components in our framework: the Depth Prior Learning module (DPL), which enriches each LiDAR point with depth priors, and the BGRF, which integrates voxel-based and point-based RoI features.

\begin{table}[htbp]
\centering
\caption{Ablation study on the KITTI validation set for the Car category using AP R40, evaluating the impact of using depth priors (DPL) and the BGRF.}
\label{tab:ablation_car}
\resizebox{\columnwidth}{!}{%
\begin{tabular}{ccccc|c}
\toprule
\textbf{DPL} & \textbf{BGRF} & \textit{Easy} (3D) & \textit{Mod}. (3D) & \textit{Hard} (3D) & \textit{Mod}. (BEV) \\
\midrule
 &         & 92.35 & 85.05 & 82.65 & 91.25 \\
\checkmark &       & 93.13 & 85.34 & 83.04 & 91.20 \\
\checkmark & \checkmark & \textbf{93.12} & \textbf{86.13} & \textbf{83.65} & \textbf{91.99} \\
\bottomrule
\end{tabular}
}
\label{tab:abla}
\end{table}

As shown in Tab. \ref{tab:abla}, introducing DPL alone yields a marginal gain (85.34 vs. 85.05) under the \textit{Mod}. setting, indicating that auxiliary geometric cues can enhance point-wise feature representation.  
When only BGRF is introduced, a more notable gain is observed (85.05 vs. 86.13), demonstrating the effectiveness of our progressive fusion mechanism in balancing global semantics and local geometry.
Combining both DPL and BGRF achieves the best performance (86.13), with improvements across all difficulty levels and a 0.74 gain in BEV AP on the \textit{Mod}. setting. 
These results validate the complementary nature of these two modules: depth priors improve the expressiveness of points, while BGRF ensures effective integration.

The significant performance gain from the BGRF module stems from three key design factors: (1) an adaptive gating mechanism that dynamically balances the contributions of global semantic context and local geometric details; (2) a bidirectional fusion strategy enabling mutual refinement between global and local features; and (3) a cascaded structure that progressively aligns and integrates RoI features across stages for deeper interaction.

\subsection{Runtime Analysis}
Our method achieves an inference speed of 9.8 FPS. A runtime breakdown shows that the PointGFE requires only 0.021s per frame, while RoI Aware Pooling adds just 0.006s. The BGRF introduces an additional 0.013s. Notably, all reported runtimes correspond to the cumulative cost of three cascaded stages, as each module is executed three times in our multi-stage refinement pipeline. 

In contrast, the RoI Grid Pooling, directly adopted from Voxel R-CNN \cite{_J_Deng___VRCTHPVb3OD___}, takes 0.044s, suggesting that this component remains a computational bottleneck.

\subsection{Qualitative Analysis}

\begin{figure*}[htbp]
\centerline{\includegraphics[width=0.80\linewidth]{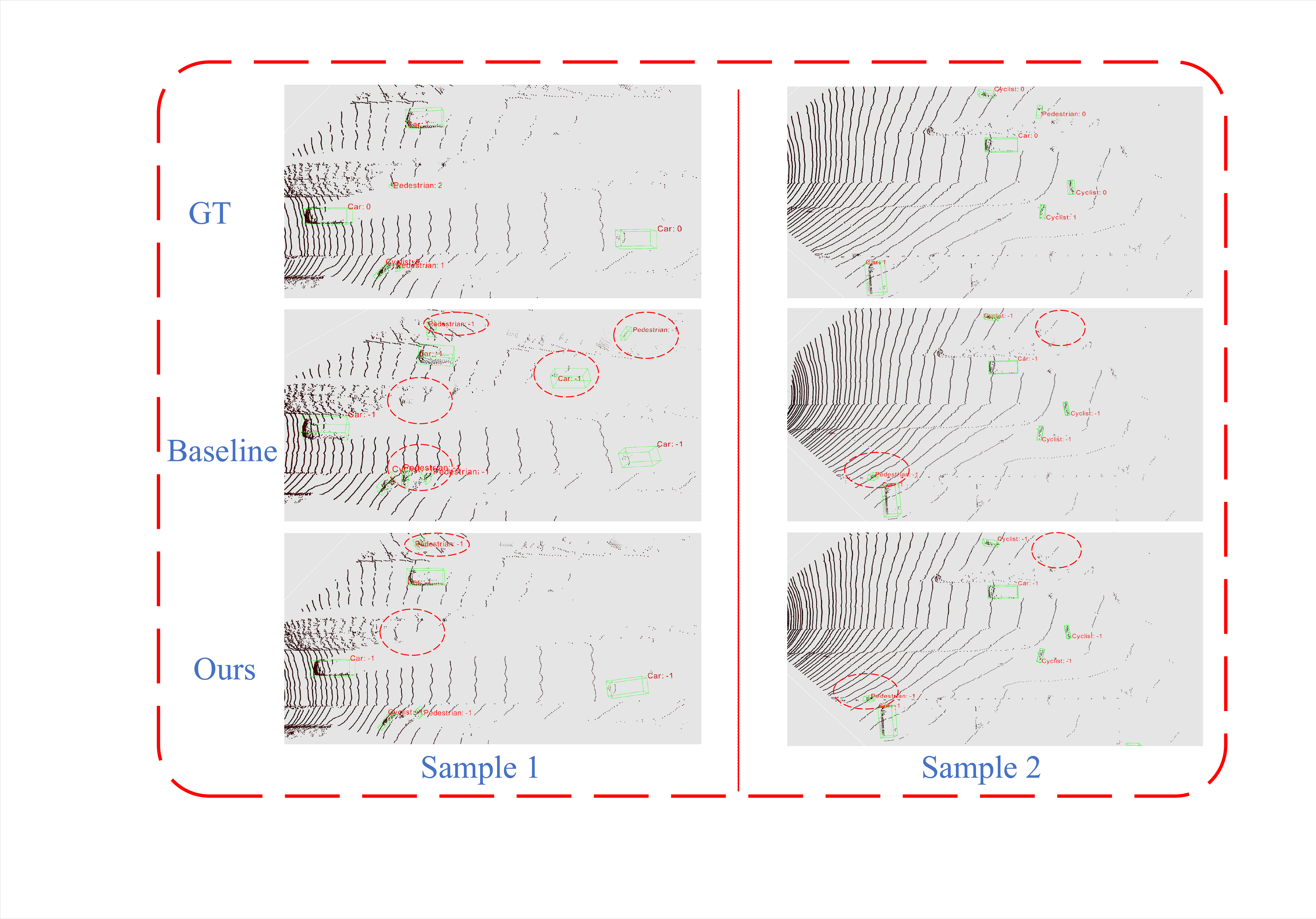}}
\caption{Qualitative comparison of 3D detection results on the KITTI validation set. Two representative scenes (Sample 1 and Sample 2) are shown in BEV. From top to bottom: ground truth, baseline predictions, and predictions from our proposed method. Red ellipses highlight regions where the baseline model produces false positives or misses objects, while our method demonstrates improved detection quality, particularly for small or partially occluded objects such as pedestrians and cyclists.}
\label{fig:ana}
\end{figure*}

To qualitatively assess the effectiveness of our proposed method, we compare detection results from the baseline and our approach on representative scenes. As illustrated in Fig. \ref{fig:ana}, each scene is visualized in BEV, showing LiDAR point clouds along with the ground truth, predictions from the baseline, and predictions from our method. Regions of interest are highlighted with red ellipses for easier comparison.

Compared with the baseline, our method produces more accurate and complete 3D bounding boxes, particularly for small or partially occluded objects such as pedestrians and cyclists. The baseline model frequently misses these targets or generates fragmented predictions in sparse or distant regions.

Nonetheless, some distant small objects remain difficult to detect accurately, indicating that further refinement may be needed in extremely sparse or occluded cases.

\section{CONCLUSION}
In this paper, we propose a 3D object detection framework that enhances LiDAR representations by integrating depth priors predicted by foundation models. These priors are fused with LiDAR attributes to form enriched point features, providing complementary geometric cues beyond raw reflectance. To leverage this information, we introduce a Dual-Path RoI feature extraction strategy and a multi-stage bidirectional gated fusion module for adaptive integration of global and local cues. Experiments on the KITTI benchmark show consistent improvements on challenging categories such as pedestrians and cyclists, validating the effectiveness of embedding geometric priors.


However, we observe a slight performance drop on the car category, likely due to misalignment between image-predicted depth and LiDAR measurements. Additionally, runtime analysis reveals that the RoI Grid Pooling module remains a computational bottleneck.


These findings highlight both the potential and limitations of integrating depth priors. Future work will explore adaptive prior integration based on object scale or scene context, and investigate alternative priors like semantics or surface normals to enhance robustness.

\normalsize
\bibliography{main}

\end{document}